\documentclass{article}

\usepackage[final]{nips_2017}

\usepackage[utf8]{inputenc} 
\usepackage[T1]{fontenc}    
\usepackage{cleveref}
\usepackage{url}            
\usepackage{booktabs}       
\usepackage{amsfonts}       
\usepackage{amsthm}
\usepackage{nicefrac}       
\usepackage{microtype}      
\usepackage{todonotes}
\usepackage{multirow}
\usepackage{algorithmic,algorithm}
\usepackage{amsmath,amssymb,amsthm}
\usepackage{tikz}
\usepackage{tikz-qtree}
\usetikzlibrary{positioning}
\usetikzlibrary{arrows}
\usepackage[tight,TABTOPCAP]{subfigure}

\usepackage{natbib}
\usepackage{etoolbox}       

\usepackage{booktabs}       
\usepackage{amsfonts}       
\usepackage{nicefrac}       
\usepackage{microtype}      

\newcommand{\R}{\mathbb{R}}

\newtheorem{remark}{Remark}

\setcitestyle{square}

\makeatletter
\patchcmd{\NAT@citex}
  {\@citea\NAT@hyper@{%
     \NAT@nmfmt{\NAT@nm}%
     \hyper@natlinkbreak{\NAT@aysep\NAT@spacechar}{\@citeb\@extra@b@citeb}%
     \NAT@date}}
  {\@citea\NAT@nmfmt{\NAT@nm}%
   \NAT@aysep\NAT@spacechar\NAT@hyper@{\NAT@date}}{}{}

\patchcmd{\NAT@citex}
  {\@citea\NAT@hyper@{%
     \NAT@nmfmt{\NAT@nm}%
     \hyper@natlinkbreak{\NAT@spacechar\NAT@@open\if*#1*\else#1\NAT@spacechar\fi}%
       {\@citeb\@extra@b@citeb}%
     \NAT@date}}
  {\@citea\NAT@nmfmt{\NAT@nm}%
   \NAT@spacechar\NAT@@open\if*#1*\else#1\NAT@spacechar\fi\NAT@hyper@{\NAT@date}}
  {}{}
\makeatother

\DeclareMathOperator*{\argmax}{arg\,max}

\title{Deep Reinforcement Learning for Dexterous Manipulation with Concept Networks}

\author{
         Aditya Gudimella \thanks{Contributed equally}, Ross Story \thanks{Contributed equally}, Matineh Shaker, Ruofan Kong,\\
          Matthew Brown, Victor Shnayder, Marcos Campos\\
  Bonsai \\
  Berkeley, CA 94704 \\
  \texttt{\{aditya.gudimella,~ross.story,~matineh.shaker,~ruofan.kong\}@bons.ai} \\
  \texttt{\{matthew.brown,~victor.shnayder,~marcos.campos\}@bons.ai} \\
}

\begin{document}

\newcommand{\cmd}[1]{\textbackslash\texttt{#1}}
\defcitealias{companion}{GMS}

\maketitle
\begin{abstract}
Deep reinforcement learning yields great results for a large array of problems, but models are generally retrained anew for each new problem to be solved. Prior learning and knowledge are difficult to incorporate when training new models, requiring increasingly longer training as problems become more complex. This is especially problematic for problems with sparse rewards. We provide a solution to these problems by introducing Concept Network Reinforcement Learning (CNRL), a framework which allows us to decompose problems using a multi-level hierarchy. Concepts in a concept network are reusable, and flexible enough to encapsulate feature extractors, skills, or other concept networks. With this hierarchical learning approach, deep reinforcement learning can be used to solve complex tasks in a modular way, through problem decomposition. We demonstrate the strength of CNRL by training a model to grasp a rectangular prism and precisely stack it on top of a cube using a gripper on a Kinova JACO arm, simulated in MuJoCo. Our experiments show that our use of hierarchy results in a 45x reduction in environment interactions compared to the state-of-the-art on this task. 
\end{abstract}

\section{Introduction}
\label{intro}
Learning goal-directed skills is a major challenge in reinforcement learning when the environment's feedback is sparse. The difficulty arises from insufficient exploration of the state space by an agent, and results in the agent not learning a robust policy or value function. The problem is further exacerbated in high-dimensional tasks, such as in robotics. Although the integration of non-linear function approximators, such as deep neural networks, with reinforcement learning has made it possible to learn patterns and abstractions over high dimensional spaces (\cite{silver2016mastering, mnih2015human}), the problem of exploration in the sparse reward regime is still a significant challenge. Rarely occurring sparse reward signals are difficult for neural networks to model, since the action sequences leading to high reward must be discovered in a much larger pool of low-reward sequences. In addition to the above difficulties, robotics tasks that involve dexterous manipulation of objects have the additional challenges of tight precision requirements, complex contact dynamics, and highly variable object geometries.

In such settings, one natural solution is for the agent to learn, plan, and represent knowledge at different levels of temporal abstraction, so that solving intermediate tasks at the right times helps achieve the final goal. \cite{sutton1999between} provided a mathematical framework for extending the notion of ``actions'' in reinforcement learning to ``options,'' which are policies taking a certain action over a period of time. The duration of execution of the option policy is specified by the time it will take the agent to meet an intermediate goal intrinsic to the option. The goal is a termination condition for the policy, defined based on the state space.

In this work, we present Concept Network Reinforcement Learning (CNRL), an industrially applicable approach to solving complex tasks using reinforcement learning, which facilitates problem decomposition, allows component reuse, simplifies reward functions, trains quickly and robustly, and produces a policy that can be executed safely and reliably when deployed. Inspired by Sutton's options framework, we introduce the notion of ``Concept Networks'' which are tree-like structures in which leaves are ``sub-concepts'' (sub-tasks), representing policies on a subset of state space. The parent (non-leaf) nodes are ``Selectors,'' which include policies on which sub-concept to choose at each time during an episode. 

Unlike the options in the framework of \cite{sutton1999between}, concepts within concept networks are not indivisible. Each concept can be a trained multi-step policy or a primitive action (like setting joint velocities to control individual fingers of a robotic arm). In addition, they can be other concept networks, creating a multi-level hierarchy, or classical controllers. Concepts can also be used for perception or other state transformation, instead of action generation. This enables further simplification of the problem each individual concept has to solve.

The flexibility of CNRL allows us to apply state transformations such as partitioning or extending the state space. This permits re-combining sub-concepts that may have been trained on different state spaces without having to retrain them. 

By treating the sub-concepts in a task as black-box components implementing entire skills, we are able to use much simpler reward functions when learning the overall task. Since the sub-tasks are much simpler than tackling the entire problem at once, their goals can often be defined on subsets of state space, significantly constraining the necessary exploration and leading to data-efficient learning even in complex environments. In addition, the approach is agnostic of the algorithms used to create and, if necessary, train a concept: each concept is treated as a black box by the rest of the concept network. This makes concepts reusable, meaning the same trained concept can be directly used in multiple concept networks. To speed up training and ensure that concepts are only executed in regions where they have been trained, each concept can include a restriction on the set states in which it can execute.

We demonstrate CNRL via the task of grasping a rectangular prism and stacking it precisely on top of a cube. This task closely reflects many problems faced in real robotics tasks, and illustrates several of the difficulties CNRL addresses. First, the problem is high dimensional. Second, the task is composed of several sub-problems, such as moving, grasping, and stacking, that are independent of one another yet common to many related tasks. This is typical of the range of real-world robotics problems. Third, the control precision required to solve this task makes it difficult to solve with a classical, hard-coded controller. Finally, for the complete task to be successful, each sub-task needs to be mastered: it would not be possible to stack a prism onto a cube if it were not grasped correctly in the first place.

To summarize, the core contributions of CNRL are as follows:

\begin{enumerate}
    \item It enables a multi-level hierarchy of problem decomposition, allowing very complex tasks to be broken down into tractable sub-problems.
    \item Conversely, it allows existing solutions to sub-problems to be composed into an overall solution without requiring re-training, regardless of the algorithms and state space definitions used to solve each sub-problem.
    \item It can simply incorporate sub-task solutions created with non-RL methods, such as classical controllers.
    \item The method of composing sub-task solutions scales to large hierarchies, significantly improving sample complexity compared to the state of the art.
    \item By limiting policy execution only to well-explored state space, it can improve the safety and reliability of execution when deployed into production.
\end{enumerate}

\section{Background}
\label{sec:background}

This section provides a brief review of the reinforcement learning (RL) problem, and how deep neural networks can be used to adapt simple Q-learning and policy optimization algorithms to tackle complex learning tasks. We focus on the algorithms used in this work, including Deep Q-Networks (DQN), Trust Region Policy Optimization (TRPO), and Hierarchical Reinforcement Learning (H-RL). We present the generic framework of these algorithms, with just the background needed to understanding our approach. 

\subsection{Reinforcement Learning}

We consider the version of the reinforcement learning problem where an agent interacts with an environment $\mathcal{E}$ in discrete timesteps. At each timestep $t$, the agent observes a state $s_t \in \R^{n}$, performs an action $a_t \in \R^{n}$, transitions to a new state $s_{t+1} \in \R^{n}$, and receives feedback reward $r_t \in \R$ from $\mathcal{E}$. The goal of reinforcement learning is to optimize the agent's action-selecting policy such that it achieves maximum expected return.

The sequence $(s_0, a_0, r_0, s_1, ..., s_t, a_t, r_t)$ is modeled as a Markov Decision Process (MDP) with state transition probability $P(s_{t+1}{|}s_t, a_t)$ and distribution over initial state $s_0$. We denote the agent policy $\pi(a_t|s_t)$ in terms of the probability distribution over the actions $a_t$, and define the return $R_t = \sum_{\tau=t}^{T}{\gamma^{(\tau-t)}r_{\tau}(s_{\tau}, a_{\tau})}$ as expected discounted reward, with the discounted factor $\gamma \in (0, 1)$, the received reward $r_{\tau}(s_{\tau}, a_{\tau})$, and the time step $T$ at the terminal state. The standard RL function $Q^{\pi}$, is the critic function for evaluating the value of each state per each action. 

\begin{equation}
\label{q_func}
Q^{\pi}(s_t, a_t) = E_{s_t, a_t, ...}[R_t|s_t, a_t]
\end{equation}

\subsection{Q-Learning and the Deep Q-Networks}
Deep Q-Network is an extended framework of the Q-Learning algorithm (\cite{watkins1992q}), with an approximation of the critic function (\ref{q_func}) using deep neural networks (\cite{mnih2015human}). Similar to Q-Learning, DQN solves the RL problem via maximizing (\ref{q_func}), in which the solution satisfies the Bellman equation:

\begin{equation}
Q^{\ast}(s_{t}, a_{t}) =E[r_t + \gamma\max\limits_{a_{t+1}}Q^{\ast}(s_{t+1}, a_{t+1})|s_t, a_t]
\end{equation}

where $a_t = \argmax Q^{\ast}$ is the greedy policy. With random initialization, a $Q$ function iteratively updated using the Bellman equation converges to the optimal solution via exploration on $s_t$ and $a_t$.  DQN approximates the $Q$ function with a neural network, with the policy converging toward the optimal solution via periodic updates to the parameters of the approximate $Q$ function. 
With DQN, the solution to the Bellman equation is achieved by solving a least-square convex optimization problem with the following loss function:

\begin{equation}\label{q_loss}
L(\theta) = E_{s_t, a_t, ...}[(r_t + \gamma\max\limits_{a_{t+1}}\tilde{Q}(s_{t+1}, a_{t+1}) - Q(s_t, a_t;\theta))^2],
\end{equation}

where $\tilde{Q}$ is the slowly-updating target Q-network that is used to periodically adjust the parameters of the $Q$-network. An experience replay buffer stores $(s_t, a_t, r_t, \tau_t, s_{t+1})$ in a data structure, with $\tau_t$ as the termination flag, and the $Q$ network is trained based on randomly sampled mini-batches from this buffer. To ensure adequate exploration of the state space, policy selection is based on an $\epsilon$-greedy strategy, that selects a random action with probability of $\epsilon$, which is annealed after each training episode.
More details and techniques on DQN can be found in (\cite{mnih2015human}) and (\cite{van2016deep}).

\subsection{Policy Optimization}
\label{policy_optim}

Policy optimization methods learn the policy directly, and adjust it to make higher rewards more likely given the observed sequences of states, actions, and rewards. There exists a long line of work in the literature, improving the robustness and scalability of such methods. These include methods based on derivative-free optimization and policy gradients.

The two main approaches based on derivative-free optimization are Cross Entropy Method (\cite{cem}) and Covariance Matrix Adaptation (\cite{cma-es}). They frame the problem as stochastic optimization, in which the distributions of policy parameters are repeatedly updated using statistics drawn from the most successful sampled paths, and the estimated distributions converge to the optimized solution. The benefits are high scalability and fast convergence on learning, but they may not perform as well as gradient based policy optimization algorithms on problems such as the game Tetris (\cite{gabillon2013approximate}).

Policy gradient methods are another active field of research in policy optimization. They mainly refer to techniques that optimize the expected return function with respect to policy parameters using gradients. The main challenge here is to approximate the gradient with high accuracy, when the reward is delayed. \cite{introtopg} introduces an approach using the Finite Difference Method, where gradient estimation is formulated as a regression problem such that policy gradients fit the temporal difference of the expected reward over small tuning in policy parameters. The weakness of the approach is it requires prior knowledge of system dynamics, as inappropriate parameter tuning leads to learning divergence. 

Tackling policy optimization via stochastic optimization, \cite{benbrahim1997biped} computes the policy gradients using the likelihood ratio, which yields fast policy convergence. However, the method has limitations on training with deterministic policies. To address this problem, \cite{dpg} and \cite{ddpg} develop a path-wise gradient method called "Deterministic Policy Gradient" that computes the policy gradients using the derivative between the output of a critic function and the policy parameters. By approximating the critic and policy functions using Neural Networks, they demonstrate successful results on numerous robotics tasks. Unfortunately, current implementations are limited to continuous action spaces, and the performance of the algorithm is sensitive to hyperparameter tuning. 

Based on the "Conservative Policy Iteration" algorithm (\cite{conserve_iter}), \cite{schulman2015trust} proposes Trust Region Policy Optimization (TRPO), an algorithm that maximizes the "monotonic improvement" term with a stochastic policy constraint, in which the policy gradient is estimated. In contrast to other policy gradient methods, TRPO improved learning stability and accuracy, as well as faster convergence speed.

For our experiments in this paper, we use the TRPO algorithm with generalized advantage estimation (\cite{schulman2015high}), as it yields more accurate training results on a wide variety of reinforcement learning tasks with little tuning on hyperparameters.

\subsection{Hierarchical Reinforcement Learning}

Effective exploration is one of the main challenges in MDPs. Although methods like $\epsilon$-greedy can be effective, in large state spaces they are insufficient to explore the full space. To tackle this problem, one can use goals and temporal abstractions: at each time $t$ and for each state $s_t$, a higher level controller chooses the goal $g_t \in G$ where $G$ is the set of all possible goals currently available for the controller to choose from. Goals provide intrinsic motivations for the agent so that it finishes the overall complex task by choosing a sequence of goals in the right order. 

Each goal remains active for some amount of time, until a predefined terminal state is reached. There is an internal critic which evaluates how close the agent is to satisfying a terminal condition of $g_t$ and sends the appropriate reward $r_g(t)$ to the controller. The objective of the controller is to maximize accumulated rewards received from the environment when the agent executes the policy defined by $g_t$. This setup is very similar to classical RL, except that an extra layer of abstraction is defined on the set of actions, so that there are specific actions for each of the goals. Different approaches to hierarchical RL result in variants on this overall approach, choosing different trade-offs in flexibility, training speed, and other properties. We describe our approach to hierarchical RL below.

\section{Related Work}
\label{sec:related-work}

\subsection{Reinforcement Learning with Temporal Abstractions}
\cite{sutton1999between} propose the options framework which extends the usual notion of action into closed-loop policies for taking actions over an extended period of time (options). This sort of temporal abstraction makes it possible to solve complex tasks by decomposing them into a combination of high level abstractions and primitive actions. The framework allows ways of changing or learning the internal structure of options, improving existing options or even combining a given set of options into a single overall policy. The theory extends Markov Decision Processes (MDPs) for reinforcement learning by building on the theory of Semi-Markov Decision Processes (SMDPs). 

A shortcoming of this approach is that the structure of an option is not readily exploited. As \cite{precup2000temporal} writes, ``SMDP methods apply to options, but only when they are treated as opaque indivisible units. Once an option has been selected, such methods require that its policy be followed until the option terminates. More interesting and potentially more powerful methods are possible by looking inside options and by altering inside their internal structure.'' Our concepts are equivalent to the options described by the framework in that a concept can be a temporally extended action or a primitive. At the same time, our concepts are not indivisible: they can be broken down into sub-concepts to lead to a truly hierarchical reinforcement learning framework.

\cite{kulkarni2016hierarchical} propose a scheme for temporal abstraction that involves simultaneously learning options and a control policy to compose options in a deep reinforcement learning framework. The authors use goals to enable better exploration of the state space. The agent focuses on learning sequences of goals in order to maximize the cumulative extrinsic reward while learning options simultaneously. The agent only uses a two stage hierarchy (consisting of a controller and a meta-controller), and is thus limited in its ability to scale with the number of goals. In contrast, our fully hierarchical approach scales to many more goals.

\cite{tessler2017deep} propose a hierarchical model that is able to retain learned knowledge and transfer the knowledge to new tasks. They achieve the knowledge transfer by using a variation of policy distillation (\cite{rusu2015policy}). They tackle the problem of scalability (with an increasing number of skills) by encapsulating multiple policies into a single distilled network, and use temporally extended actions to solve tasks with lower sample complexity.

\subsection{Applications in Robotics}
\label{robotics_tasks}

\cite{levine2016visuo} use Convolutional Neural Networks (CNNs) to support vision-based robot control (visual servoing), training a network to predict the probability of success of a given grasp attempt by a classical controller using camera images, and aborting and re-planning if the success probability is too low. They used between 6 and 14 robots to gather samples in parallel, demonstrating the value of parallel sampling when applying deep learning to robotic control. This approach requires an already existing controller that can accomplish the task reasonably well, and augments it with deep learning.

\cite{gu2016anaf} go further to learn full control policies capable of grasping, picking and placing, and door opening, in simulation and on real platforms. They use staged reward functions, shaping, and parallel sampling in simulation to train policies capable of completing complex tasks from low dimensional input in 500,000 samples. Though they break their reward functions for complex tasks into stages, these stages are not recomposable.

\cite{popov2017data} tackle the problem of combining a set of options into a single more complex overall policy by using a form of apprenticeship learning that facilitates learning of long-horizon tasks even with sparse rewards. They also introduce a recipe for constructing reward functions for complex tasks consisting of a sequence of sub-tasks. Using previously trained policies for sub-tasks they sample states from successful policy executions and use those as initial states when training a policy to complete the full task. This biases exploration towards regions of state space useful for solving the sub-concepts, accelerating policy convergence. Using these pre-trained sub-concepts for exploration, in conjunction with staged shaped rewards, they were able to learn the task we explore here, picking up an object and precisely stacking it on another. The optimized algorithm learned this task using approximately 1,000,000 environment transitions. Our approach for this problem, described in detail below, is able to use the solutions to sub-tasks directly, and solves the same task using only 22,000 environment transitions. 

\cite{finn2016inverse} explore an alternative method of simplifying reward function construction for complex tasks by using inverse reinforcement learning to learn a reward function from expert policy execution. By observing successful execution of a task off policy, they estimate the reward function that led to that policy based on the features provided, and then train a policy to optimize this new reward function. This has the advantage of obviating the requirement for hand crafting a reward function if an expert can demonstrate their skill using the robot. This could be combined with our approach to generate rewards for reusable sub-concepts.

\section{Concept Network Reinforcement Learning}
\label{sec:cnrl}

   \begin{figure}
        \centering
        \subfigure{
             \includegraphics[width=0.45\textwidth]{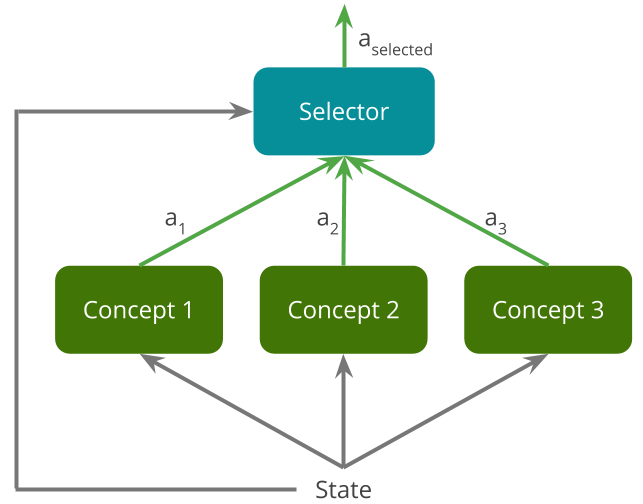}
            \label{fig:cnrl-selector}
        }
        \subfigure{
            \includegraphics[width=0.45\textwidth]{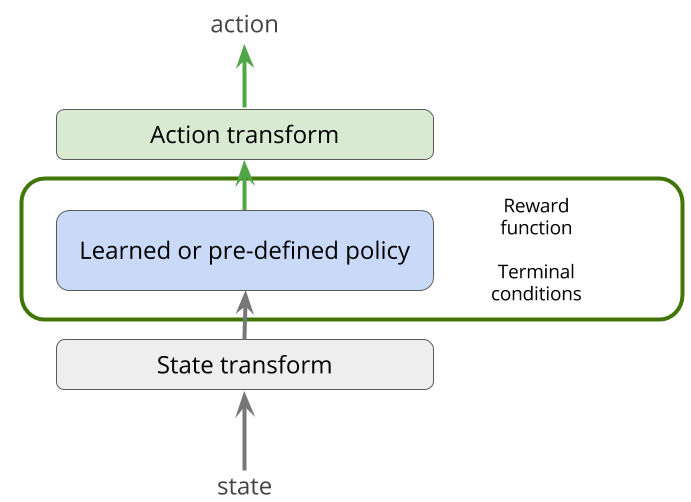}
            \label{fig:cnrl-policy}
        }
       
        \caption{(a) Selector concept. Once chosen, a child concept executes until it reaches a terminal condition.  (b) Control concepts, including (optional) state and action transformations.The reward function and terminal conditions for a state can be written in terms of the concept's transformed state, and are independent of the rest of the concept network. Of course, only learned control concepts need a reward function.}
        \label{fig:cnrl-model}
    \end{figure}       

  \begin{figure}
        \centering
        
            \includegraphics[width=0.6\textwidth]{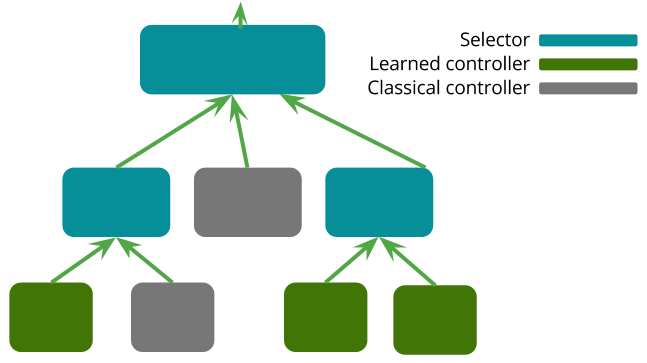}
            \label{fig:cnrl-example}
            \caption{Example of a hierarchical concept network, including three selectors, three control concepts, and two classical controllers. Each node also implicitly takes the state as input, and can be paired with input and output transformations as shown in Figure~\ref{fig:cnrl-policy}.}
  \end{figure}

As discussed earlier, an industrially applicable approach to solving complex tasks using reinforcement learning should facilitate problem decomposition, simplify reward function design, train quickly and robustly, and produce a policy that can be executed safely and reliably when deployed. Concept Network Reinforcement Learning (CNRL) is a variant of Sutton’s options framework that achieves these goals. This section describes CNRL in detail, and discusses its benefits and trade-offs.

CNRL is based on decomposing the overall learning problem into \emph{concepts}, each of which represents some aspect of the solution. There are three types of concepts: control concepts, which define actions to take in certain situations, selector concepts, which choose one of their sub-concepts to act next, and transformation concepts, which transform low-level state input into higher level perceptual features that are more useful for subsequent concepts. The overall concept network is a directed acyclic graph, with the overall system state coming in, a mixture of \emph{control}, \emph{selector}, and \emph{transformation} nodes processing that state, ultimately producing the action to execute in the environment. An example is shown in Figure~\ref{fig:cnrl-example}. We now describe each type of concept in more detail.

\subsection{Selector Concepts}

Figure~\ref{fig:cnrl-selector} shows the structure of a selector concept: selectors accept the state from the environment and choose one of a set of child concepts, which can be either other selectors or control concepts. This child concept's policy then interacts with the environment, receiving states and generating actions to transition the environment until the child reaches its terminal condition, at which point the selector again receives the new state and the new value of its own reward function, and makes a new choice. Execution can recursively descend through several selectors in turn before a control concept is reached. Treating skills implemented by child nodes as discrete units for the selector speeds exploration and avoids unnecessary backtracking---if a new child concept were selected for each time step, one concept's policy could undo the progress made by another. Our results in Section~\ref{sec:results} demonstrate how this greatly speeds up training.

The selector's children are treated as a policy-implementing black box. This allows incorporating control concepts implemented via non-RL methods such as traditional control, and allows selectors to be nested: simple skills may be grouped together under a selector to form a more complex skill, which may in turn be selected by its parent.

Selectors are trained using a discrete action algorithm -- we use DQN in our implementation. Because the chosen policies are trained separately and treated as a black box that executes a policy to termination, reward functions for selectors can be simple, typically rewarding progress toward an overall goal. If the selector's task can be solved with a small number of child policy executions, or the right child to pick is easy to deduce from the state, this becomes a simple and short-horizon reinforcement learning problem, and the selector is very quick to train. Section~\ref{sec:results} demonstrates this on the pick-and-place robotics task.

\subsection{Control Concepts}

As illustrated in Figure~\ref{fig:cnrl-policy}, a control concept takes state as input and produces an action, which can be a single-step or multi-step policy. Control concepts can be learned using RL, can use a manually coded controller, e.g. using inverse kinematics, or can be implemented using a pre-trained neural network based controller, perhaps re-used from another concept network. A policy can even learn behavior on only a part of the action space, with a following transformation node adding hard-coded behavior on the rest: for example, when learning to orient the gripper in our robotics task, we hard code that the gripper fingers should be open, and let the network learn to control the other arm joints. 

The black-box nature of concepts allows each to be specified with the most appropriate state space and action space for the task. Transformation nodes before the concept node can convert the input to an appropriate form, e.g. by omitting irrelevant elements or augmenting with derived properties. The learning problem is solved with the transformed input and an appropriate action space, and a following transformation node transforms the output as needed for following nodes.

Each learned control concept has its own reward function, independent of the overall problem. Thus, reward shaping considerations are encapsulated within concepts, and only need to be defined on the relevant portions of the concept's state and action space. Each learned control concept can also be trained with the most appropriate learning algorithm for that task. This ability to customize the training approach for each sub-problem speeds up task design and iteration, and can significantly speed up training. 

\subsubsection{Validity regions for control concepts}

Once a control concept is selected, it continues to execute its policy until it hits one of its execution terminal conditions. There are three types of terminal conditions. The first is completing the overall task, successfully ending the episode.  The second is completing the concept's task and returning control to its parent selector. The third is the system state leaving the concept's \emph{validity region}, a configurable set of states where the concept is allowed to run. The validity region also constrains where the concept may start execution. When the state is outside this region, parent selectors are not allowed to select this concept (we implement this in our system by having the control concept return a no-op action if chosen in a terminal region, so the selector learns not to make such choices. This could also be implemented by directly masking out ineligible children in the DQN output.) 

By cutting off unpromising exploration quickly, validity regions can drastically reduce the time needed to learn concepts. At deployment time, the validity region ensures that the concept can only execute its policy in regions of state space it has explored during training and in which its policy has been well characterized and deemed safe.  Without such constraints, the undefined behaviour of RL-based control policies outside the state space they have explored during training can pose a significant safety hazard when deployed into production. The validity region can be configured differently during training and deployment. This can be used to provide an additional margin for error, or further restrict the work space where execution of a given skill is permitted.

\subsection{Transformations}

Concept networks can include transformations, which can act on states or actions to adapt them for downstream use. Just like control concepts, transformations can be hard-coded, pre-trained, or learned. The two primary uses are transforming states into higher level representations and adjusting state inputs and action outputs to match each concept's requirements. Typical examples of the former include perception tasks, such as converting visual input into an object identification or converting text into a topic vector and a sentiment estimate. When such perception transformations are learned using neural networks, it may be more effective to output an embedding vector derived from the penultimate network layer, rather than the output value used for training (e.g. the object class, or the selected text topic).

Other uses of transformations include filtering out or converting state information as shown in Figure~\ref{fig:cnrl-policy}, adding in hard-coded action elements, mapping data representations into other forms (e.g. converting from cartesian to polar coordinates), and concatenating action elements from concepts controlling individual aspects of the overall task into a complete action vector. The ability to include transformations in concept networks naturally follows from the fact that each concept can be trained separately, and gives great practical flexibility in reuse and network organization. 

\subsection{Discussion}

The CNRL approach has many benefits and directions for extension. Perhaps the greatest benefit is the ability to truly decompose reinforcement learning problems into independent parts. This is crucial for applying RL to real industrial problems, allowing teams to divide and conquer: different groups can independently work on different aspects of a learning problem, quickly assemble them into a full solution, and upgrade individual components later.

Because a selector treats its child concepts as black boxes, each can be implemented using the technique most appropriate to the problem. In robotics, for example, complex tasks requiring dexterous manipulation like grasping may be implemented with deep reinforcement learning, while well-characterized tasks like moving between work spaces can be handled by inverse kinematics. For the same reason, entire concept networks are re-usable and composable: a solution to one problem can be used as a component in a larger problem.

Individual concepts in a network can be easily replaced with alternate implementations, allowing easy experimentation and incremental improvement -- a hard-coded controller can be replaced with a learned one, or an intractable concept can be further subdivided, all without requiring any change in the rest of the concept network. Additionally, the independence among all children of a selector typically allows them to be trained in parallel.

CNRL, like other hierarchical methods, makes the model more explainable than monolithic training methods: seeing the concepts activated by each selector gives higher-level insight into the behavior than simply seeing the low-level actions at each time step.

CNRL can have several extensions beyond what is described here. As presented, the task must be broken down into concepts that completely cover the state space -- if there are situations where none of the children of a selector can make progress, the overall problem will be unsolvable. We have designed and implemented a more advanced selector that can synthesize a policy to cover such gaps, and will report the details in a future publication. In certain tasks with sequential concepts, completely independent parallel training of concepts may be impossible, if the starting conditions of one concept depend on the end state of the previous concept. In such settings, the system needs to coordinate these end and start conditions among concepts. Finally, decomposing the problem into completely independent pieces prevents the system from adjusting all the elements end-to-end: each concept is trained independently. In settings where joint training or fine-tuning is crucial, one can start with a combined concept integrating several components, then split them apart again after training for reuse and independent evolution.

\section{Solving ``Grasp and Stack'' with CNRL}
\label{sec:grasp-and-stack}

\begin{figure}
    \begin{tabular}{cccc}
        \includegraphics[height=0.3\textwidth]{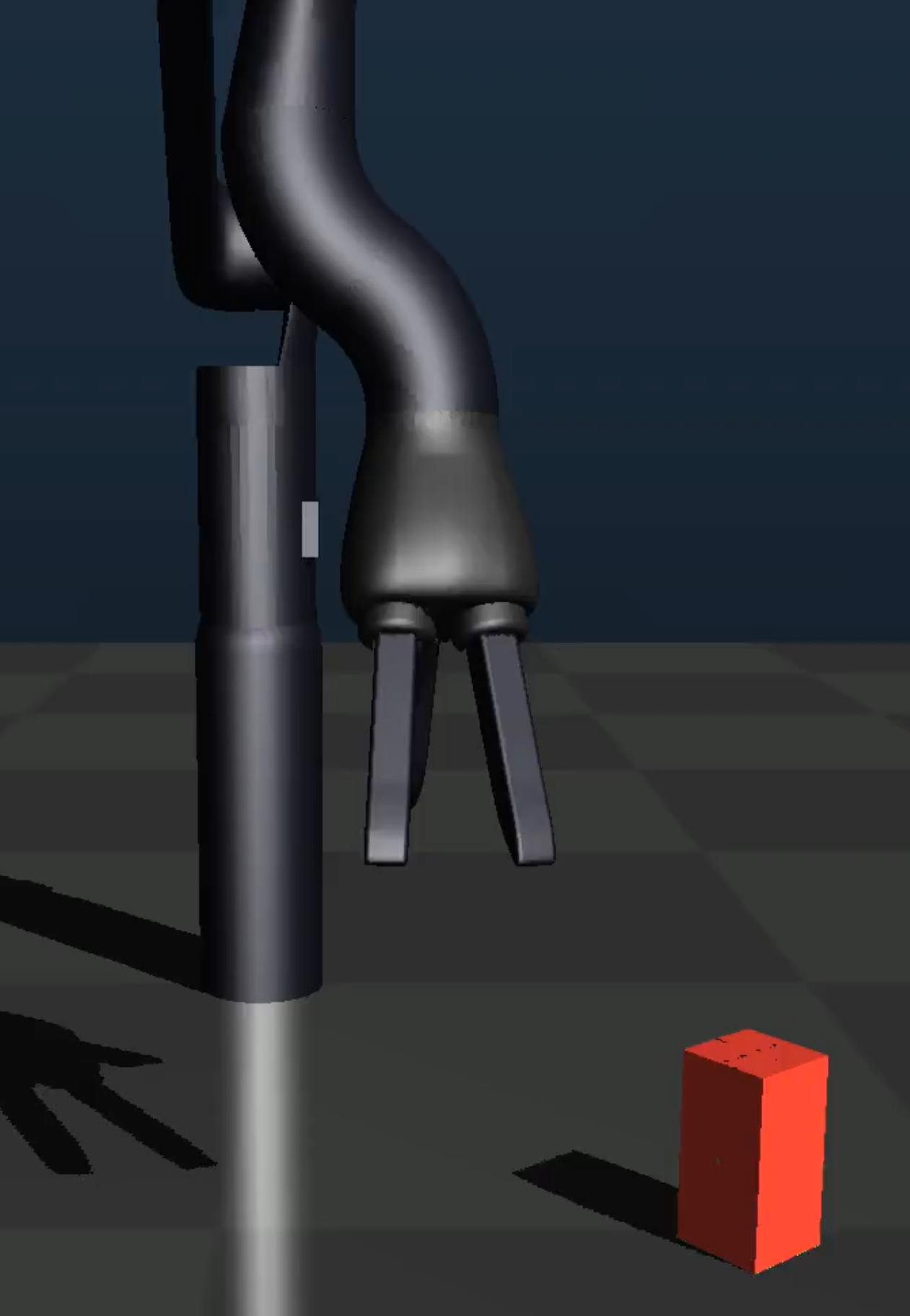} &
        \includegraphics[height=0.3\textwidth]{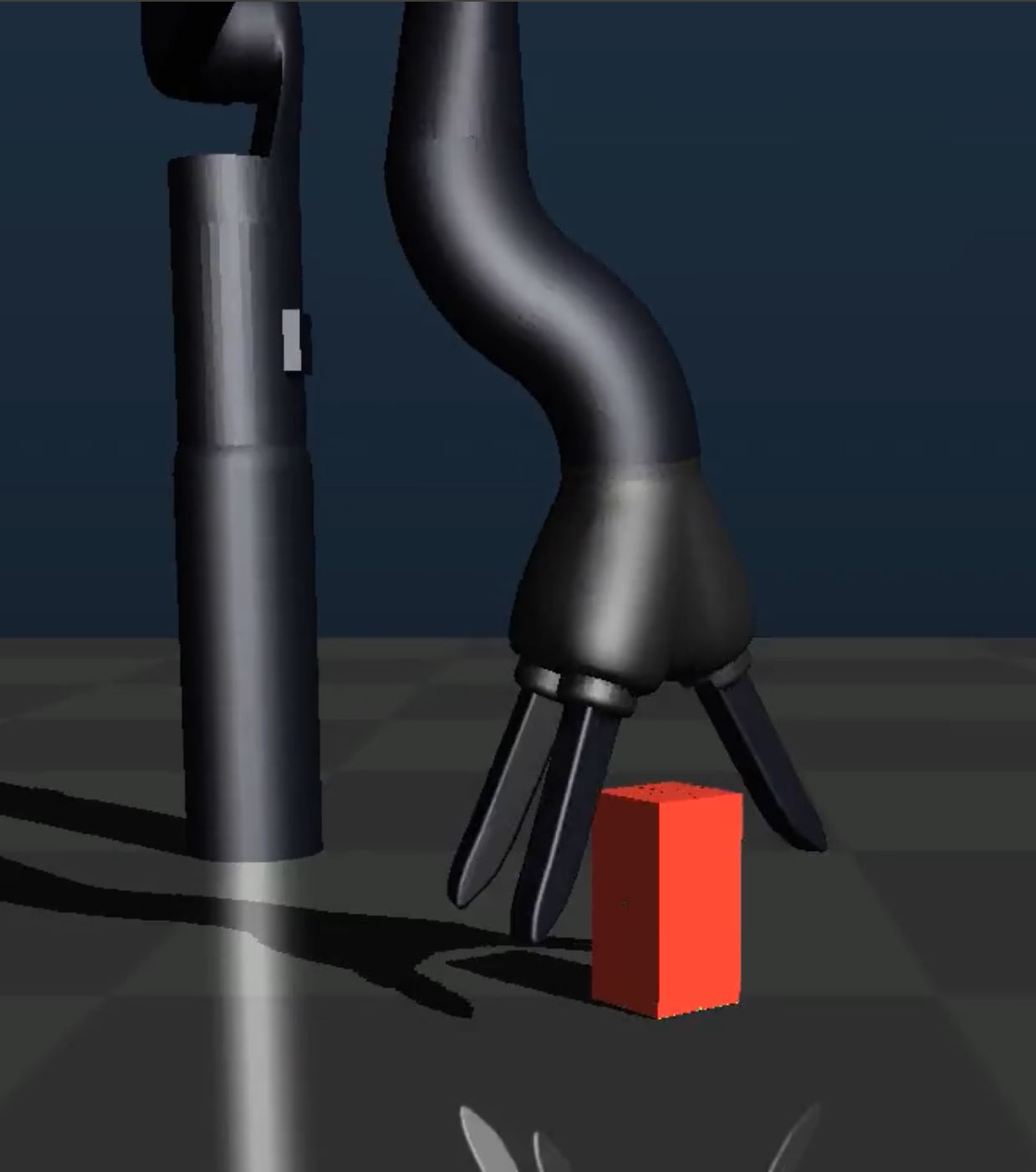} &
        \includegraphics[height=0.3\textwidth]{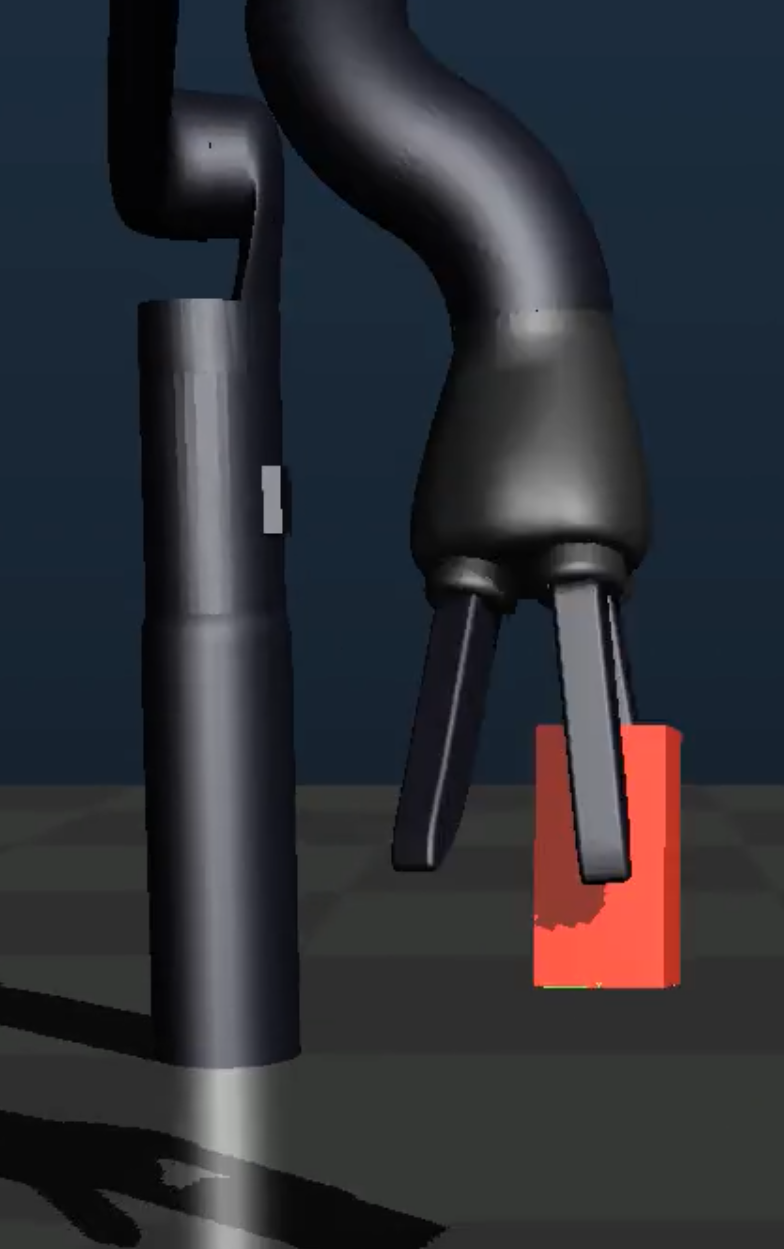} &
        \includegraphics[height=0.3\textwidth]{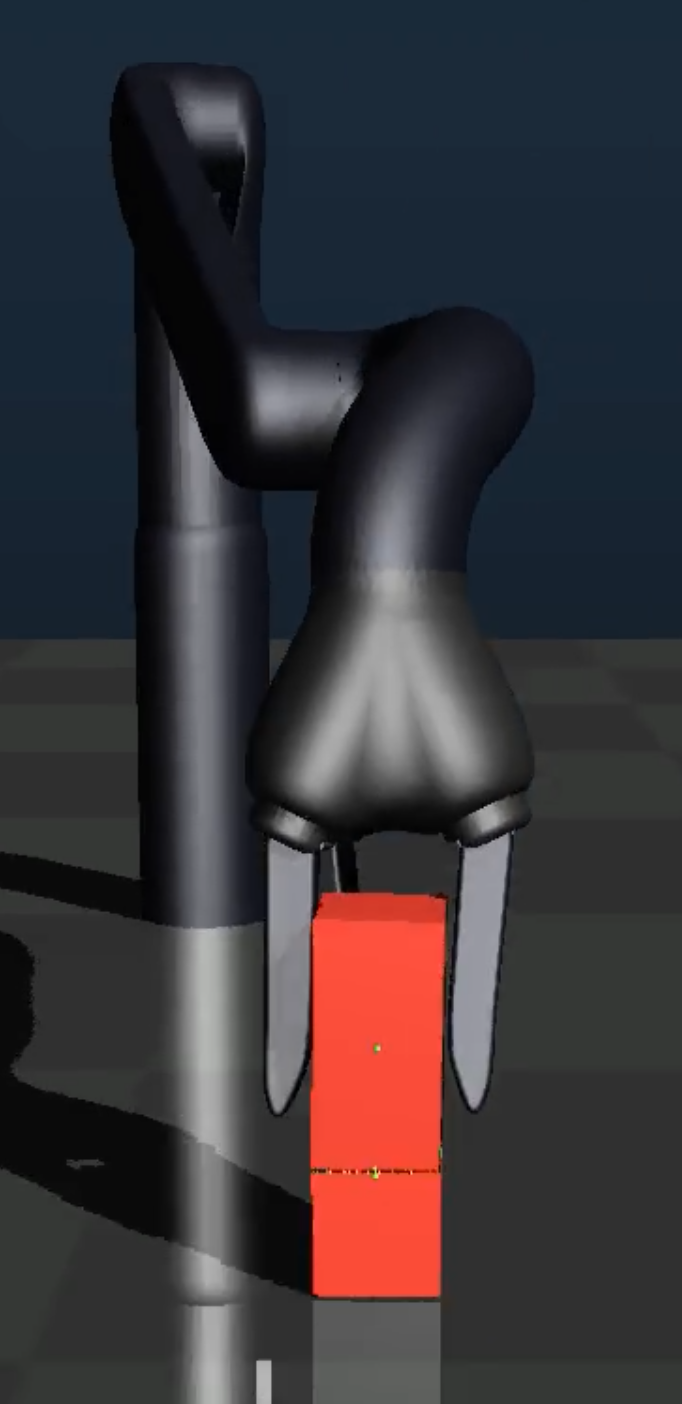} \\
    \end{tabular}
    \caption{Stages of the task from left to right: (a) Grasp staging, (b) Orient, (c) Lift, (d) Stack}
    \label{fig:stages}
\end{figure}

\subsection{Concept Network}
We demonstrate CNRL on the task of grasping a rectangular prism and precisely stacking it on top of a cube. The rectangular prism was chosen for ease of manipulation by the gripper provided on the JACO model. We initially broke the overall task down into four sub-concepts: 1) reaching the working area, 2) grasping the prism, 3) moving to the second working area, and 4) stacking the prism on top of the cube. During training we found that TRPO, the algorithm we chose for training control concepts, had difficulty learning a single policy to grasp the prism. To simplify the learning problem, we broke grasping into two sub-concepts: orienting the hand around the prism in preparation for grasping, and lifting the prism, for a total of five control concepts in the concept network. Three of these -- orienting, lifting, and stacking -- used TRPO to train, while moving to the working area for both grasp and stack (Staging-1 and Staging-2 in Fig.~\ref{fig:flat_concept_graph} and Fig.~\ref{fig:ml_concept_graph}) were handled with inverse kinematics.

We explored two concept hierarchies: a single selector with five children (Fig.~\ref{fig:flat_concept_graph}), and a multi-level tree with two selectors, one with four children and the other with two (Fig.~\ref{fig:ml_concept_graph}). In this small example there is little benefit to nesting selectors in this way, but as the size of the tree scales, the ability to encapsulate sections of the problem and separately learn the correct circumstances in which to invoke concepts to solve that subproblem will improve training parallelization, help keep large concept trees organized, and bound the complexity of the task any single selector must learn.

\subsection{State Spaces, Action Spaces, and Rewards}

The state vector provided to the agent varied from concept to concept, as did the action space. For example, the reach and grasp stages only need to consider the location of the prism, not the cube, so the cube's location was omitted from the state space. All actions correspond to target velocities for nine associated joints. In some cases, only some of the target velocities were learned, with others hard-coded---keeping the gripper always open during orient, and always closed during moving and stacking.The action and state vectors are described in Table \ref{tab:states} and \ref{tab:actions}, while the rewards and terminals are described in the appendices \ref{app:shaping} and \ref{app:terminal}, respectively.

\subsection{Experimental Setup}

The agent controlled a Kinova JACO arm simulated in MuJoCo. Episodes of the full task were terminated after 150 steps, with a control frequency of $20 \: \text{Hz}$. Subconcepts were terminated after 50 steps. All concepts were terminated early under conditions laid out in Appendix~\ref{app:terminal} to curtail unnecessary exploration and improve sample efficiency.

The positions and initial angular velocities of the arm joints started with a small random variation, and the positions of the prism and cube varied by $\pm{10}\:\text{cm}$ in the x and y axes, while the orientations varied by $\pm{1}$ radian.

\begin{figure}
\begin{center}
\begin{tikzpicture}[->,>=stealth',shorten >=1pt,auto,node distance=3cm,
                    main node/.style={draw,font=\sffamily}]

  \node[main node] (1) {Selector};
  \node[main node] (3) [below left  of=1] {Grasp};
  \node[main node] (2) [left of=3] {Staging-1};
  \node[main node] (4) [below right of=1] {Staging-2};
  \node[main node] (5) [right of=4] {Stack};
  \node[main node] (6) [below left  of=3] {Orient};
  \node[main node] (7) [below right of=3] {Lift};

  \path[every node/.style={font=\sffamily\small}]
    (1) edge [left] node [left] {} (2)
        edge node [left] {} (3)
        edge node [left] {} (4)
        edge [right] node[left] {} (5)
    (3) edge [left] node[left] {} (6)
        edge [left] node[left] {} (7);

\end{tikzpicture}
\caption{Multi-Level Concept Network}
\label{fig:ml_concept_graph} 
\end{center}
\end{figure}
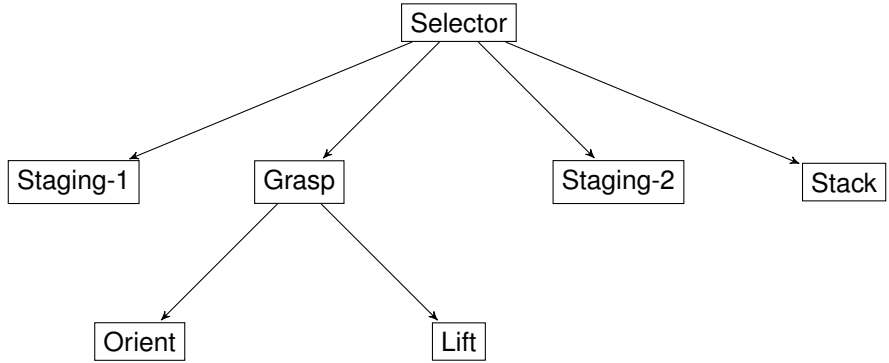

We report results using the hierarchical concept network shown in Fig.~\ref{fig:ml_concept_graph}. Stack, Orient, and Lift are control concepts trained using TRPO, while the full concept selector and the Grasp selector were trained using DQN. Each node was trained after all of its sub-concepts had finished training and their weights were frozen.

We trained the TRPO concepts using the publicly available OpenAI Baselines parallel TRPO implementation, using the ADAM optimizer and 16 parallel workers. We used default hyperparameters, including a batch size of 1024, a maximum KL divergence of 0.01, a gamma of 0.99, and a step size of 1e-3. We made no modifications to the underlying algorithm to facilitate replication and comparison.

We trained the DQN concepts using the OpenAI Baselines DQN implementation, with the ADAM optimizer and only a single worker. DQN was trained with a batch size of 64, learner memory capacity of 50000 samples, a minimum learner memory threshold of 1000 samples, an exploration probability that decayed from 1 to 0.02 over 10000 steps, a gamma of 0.98, and a learning rate of 5e-4.

\begin{figure}
\begin{center}
\begin{tikzpicture}[->,>=stealth',shorten >=1pt,auto,node distance=2.5cm,
                    main node/.style={draw,font=\sffamily}]

  \node[main node] (1) {Selector};
  \node[main node] (3) [below left of=1] {Orient};
  \node[main node] (6) [right of=3] {Lift};
  \node[main node] (2) [left of=3] {Staging-1};
  \node[main node] (4) [right of=6] {Staging-2};
  \node[main node] (5) [right of=4] {Stack};

  \path[every node/.style={font=\sffamily\small}]
    (1) edge [left] node [left] {} (2)
        edge node [left] {} (3)
        edge node [left] {} (4)
        edge [right] node[left] {} (5)
        edge [left] node[left] {} (6);

\end{tikzpicture}
\caption{Single Level Concept Network}
\label{fig:flat_concept_graph} 
\end{center}
\end{figure}
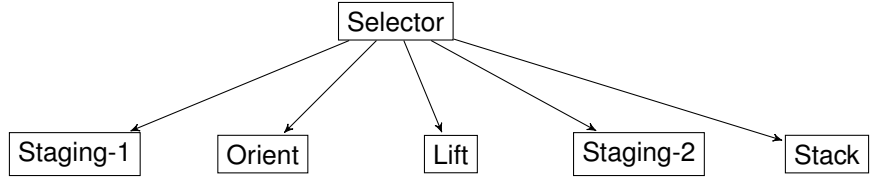

\section{Experimental Results}
\label{sec:results}

Training performance for DQN was evaluated with ten testing episodes for every 50 training episodes, with mean performance in each testing pass plotted in the selector performance graphs below. Training performance for TRPO uses the raw training episode returns, which are less representative of true policy performance but served well enough to show when the policy had converged. In plots showing the performance of DQN, the x axis represents transitions sampled so far, and the y axis represents mean episode reward. Final evaluation of robustness for both DQN and TRPO was done without exploration.

\begin{figure}
        \centering
        \subfigure{
            \includegraphics[width=.7\textwidth]{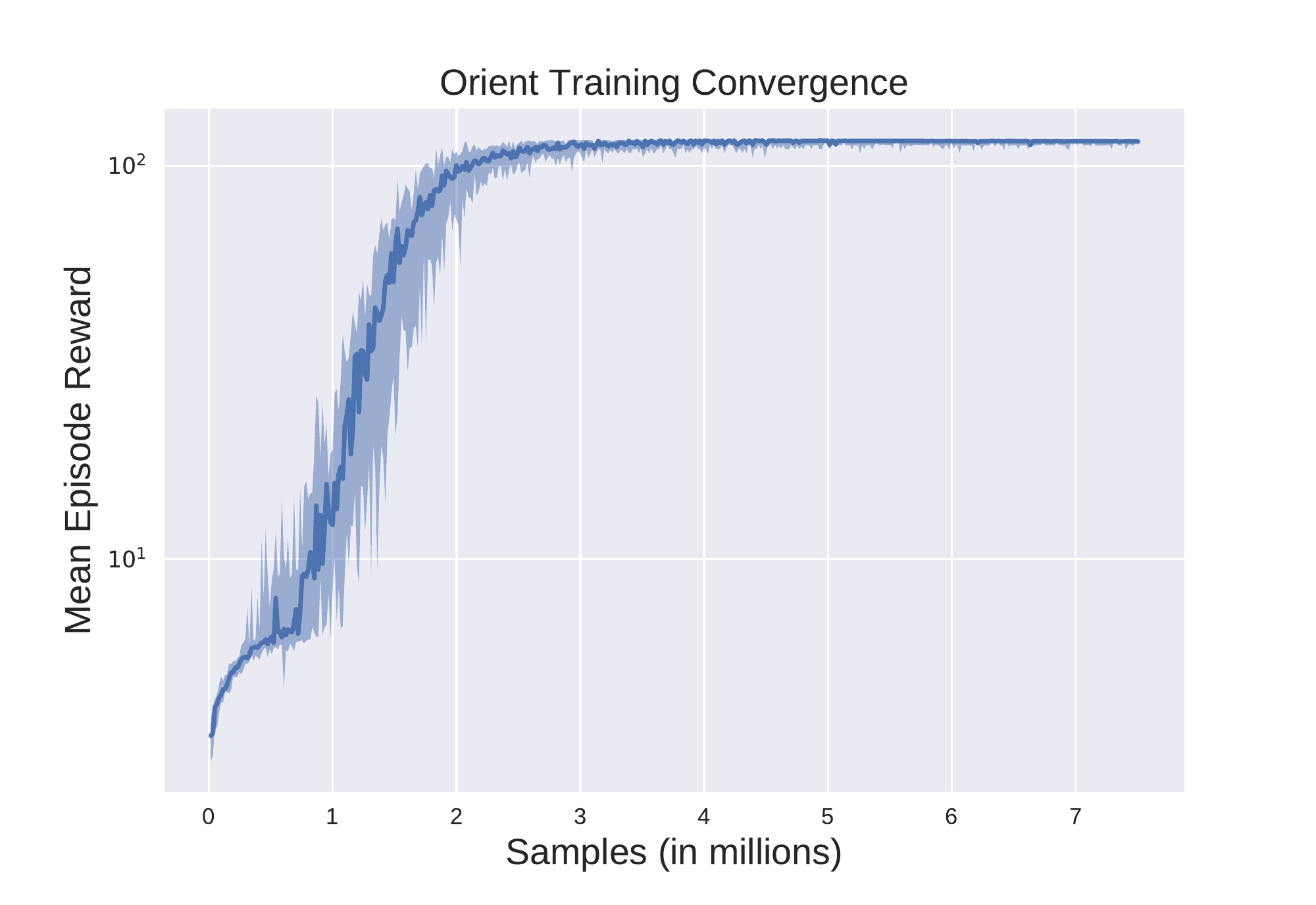}
            \label{fig:orient_training}
        }
        \subfigure{
            \includegraphics[width=.7\textwidth]{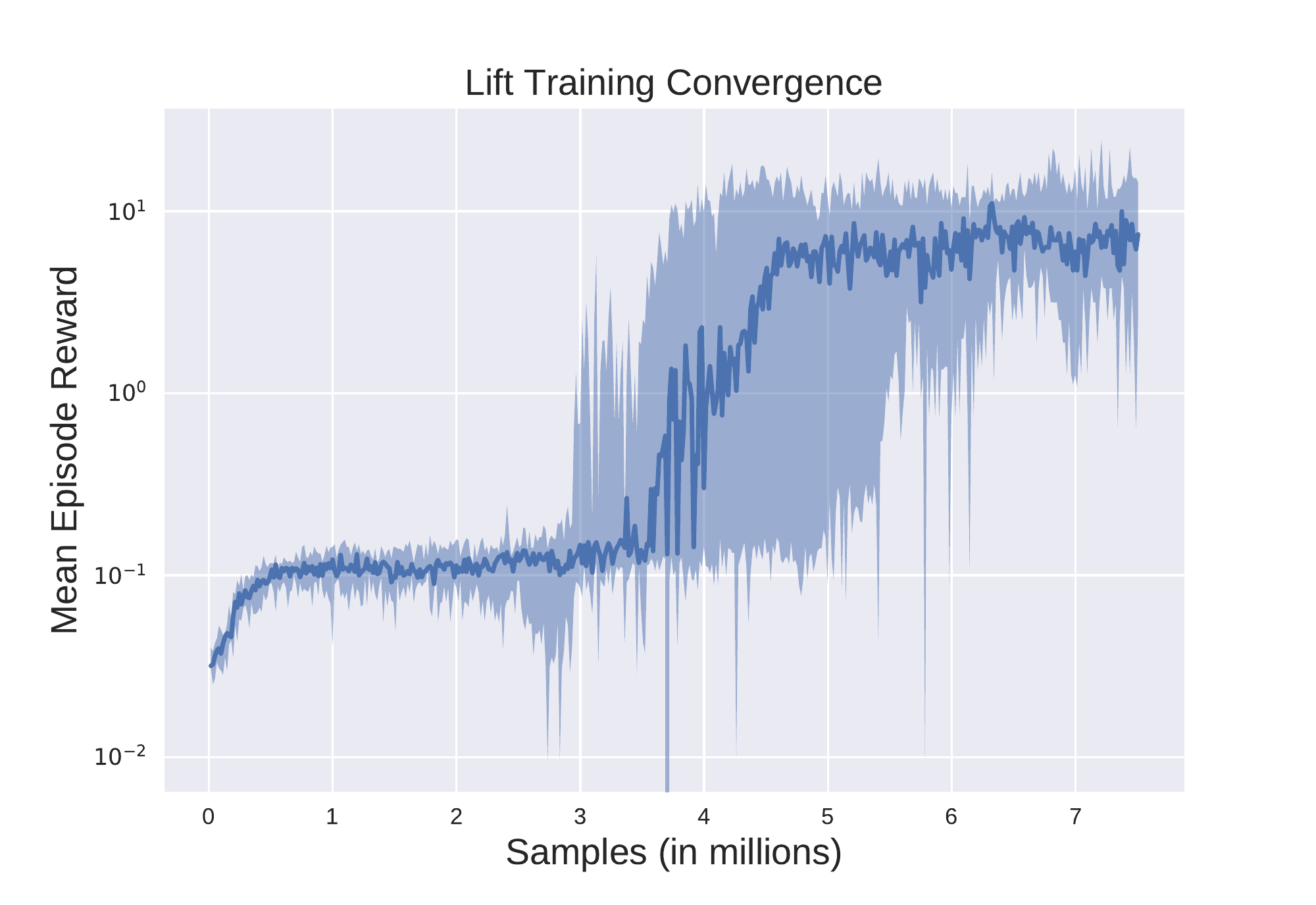}
            \label{fig:lift_training}
        }
        \subfigure{
            \includegraphics[width=.7\textwidth]{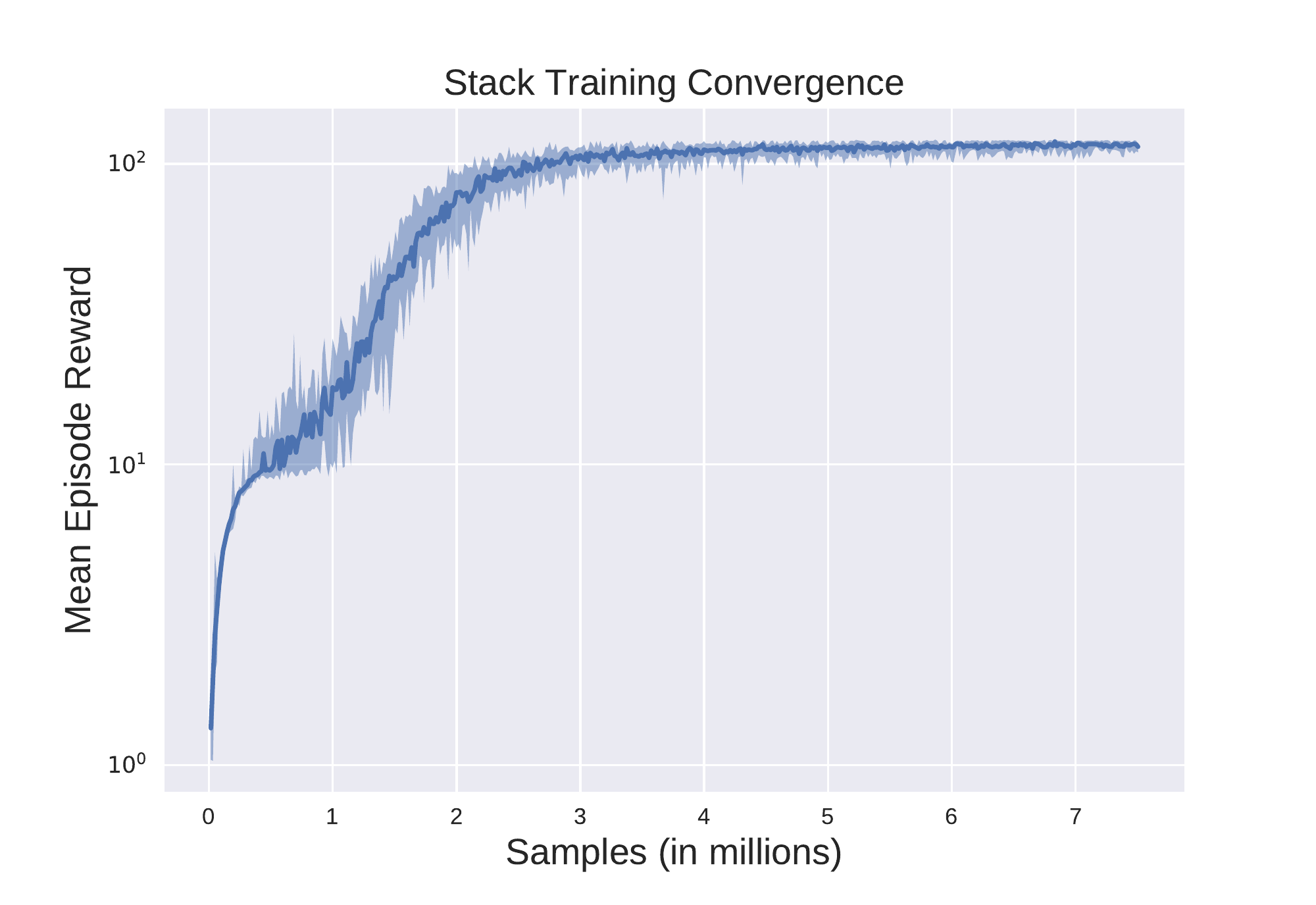}
            \label{fig:stack_training}
        }
        \caption{Mean episode return of the "Orient" (top), "Lift" (middle) and "Stack" (bottom) concepts as a function of samples seen for 10 runs. Shaded area represents the min to max; For the Lift concept, tight terminal conditions are set to encourage precise vertical lift, which makes finding a good policy more challenging; in one omitted case the agent failed to find an effective policy inside 7.5M samples.}
        \label{fig:stages}
    \end{figure}

The orient and stack concepts trained in approximately 2-3 million samples using shaping rewards and guiding terminals, without the need for hyperparameter tuning. The training graphs for the TRPO concepts are presented in Figs.~\ref{fig:orient_training}, \ref{fig:stack_training}, and \ref{fig:lift_training}. In one of the training runs the lift concept did not converge within 7.5 million samples and this data was omitted. The very tight terminal constraint on the distance the prism can move from its starting xy coordinates, designed to encourage a straight vertical lift, also increased the number of samples required to find a good policy through exploration, and in at least one case increased it beyond the sample budget we allotted. Better designed terminal conditions and rewards could undoubtedly speed up training on this task, but almost all of the policies were sufficient to effectively complete the task.

\begin{figure}
  \centering
    \includegraphics[width=0.7\textwidth]{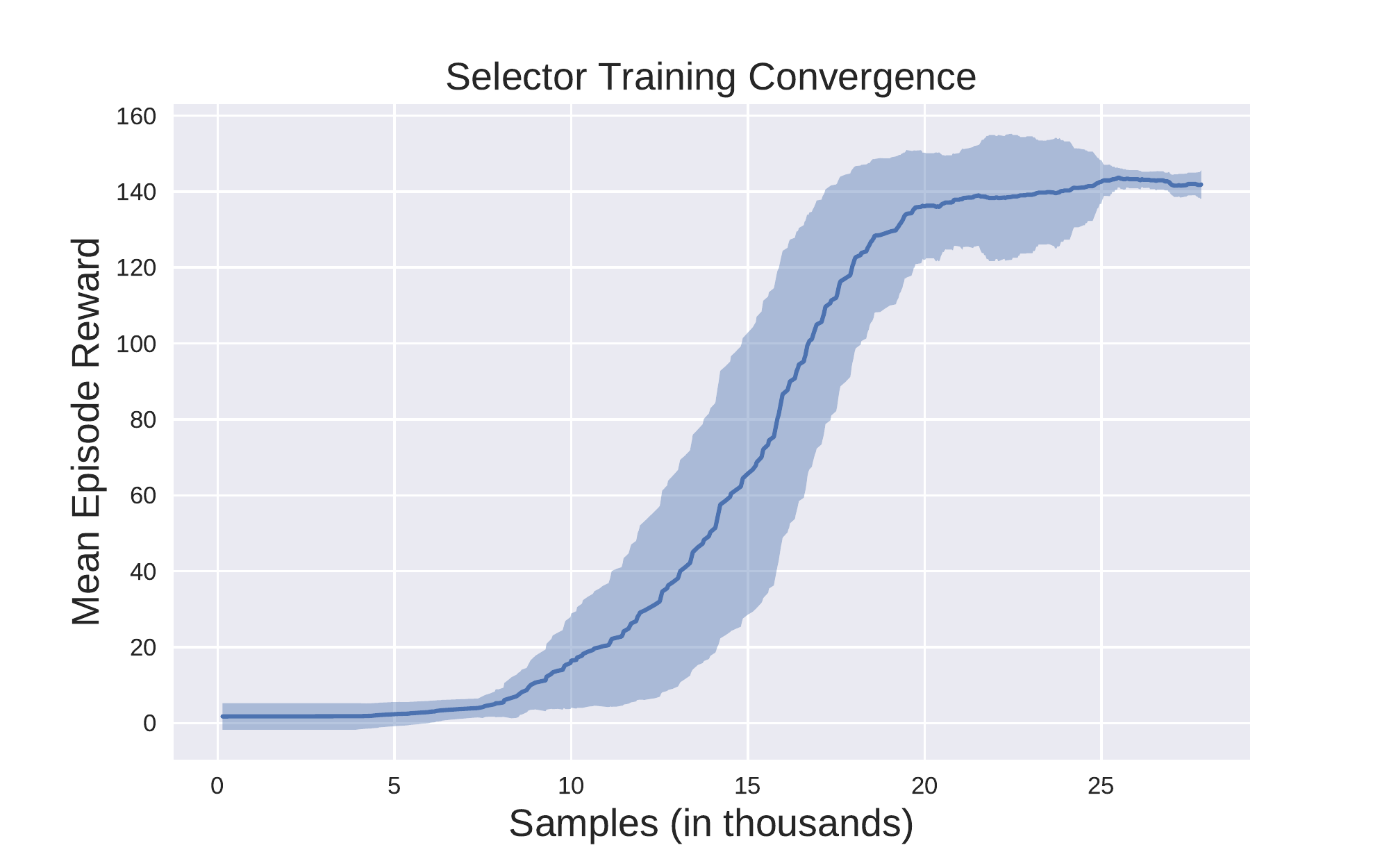}
  \caption{Mean episode return of the selector as a function of samples seen for ten runs. The shaded area is a 95\% confidence interval for the mean. Models converged on good solutions between $16 000$ and $25 000$ samples.}
  \label{fig:selector_training}
\end{figure}

The full concept selector trained in 22,000 samples (Fig.~\ref{fig:selector_training}), though the selector itself only saw 6,000 samples as it does not receive state transitions during long running execution of children. When concepts are compatible --- i.e. a concept ends within the operating constraints of another --- and there exists some chain of compatible concepts that will achieve a goal, the selector can learn to order these concepts very quickly, without the need to train a monolithic network to subsume the components. The task of ordering the concepts can be learned nearly two orders of magnitude faster than the individual concepts, or 45x faster than the single policy trained by \cite{popov2017data} using one million samples and previously trained sub-concepts.

In 500 episodes we observed no task failures during execution, both with the subconcepts executed individually in their own environments and the tree with selectors solving the full task. The concept network is able to very reliably grasp an object and precisely stack it on another, both with varying position and orientation. Videos of the trained policies can be seen at \url{http://bns.ai/robotics_blog}.

\begin{table}[htbp]
    \caption{State vectors for each of the three concepts plus selectors.}
    \label{tab:states}
    \centering
    \begin{tabular}{l|l}
        \hline
        \textbf{Concept} & \textbf{State} \\ 
        \hline
        \multirow{9}{*}{Orient} & For the six joints excluding the fingers: \\ 
                                & 1. The sine and cosine of the angles of the joints. \\
                                & 2. The angular velocities of the joints. \\
                                & 3. The position and quaternion orientation of the prism. \\
                                & 4. The orientation of the line between two of the opposed fingers \\
                                &       ~~~~     in degrees normalized by $90^\circ$. \\
                                & 5. The euclidean distance between the pinch point of the opposed \\ 
                                &       ~~~~     fingers and a point $1.5 \text{cm}$ above the prism. \\
                                & 6. The vector between the same two points above. \\
        \hline
        \multirow{6}{*}{Lift} & For all nine joints: \\ 
                                & 1. The sine and cosine of the angles of the joints. \\
                                & 2. The angular velocities of the joints. \\
                                & 3. The euclidean distance between the pinch point of the \\ 
                                &       ~~~~     opposed fingers and the center of the prism. \\
                                & 4. The vector between the same two points above. \\
                                & 5. The xy vector between the center of the prism and the \\
                                &       ~~~~     starting position of the prism. \\
        \hline                       
        \multirow{7}{*}{Stack} & For the six joints excluding the fingers: \\ 
                                & 1. The sine and cosine of the angles of the joints. \\
                                & 2. The angular velocities of the joints. \\
                                & 3. The position and quaternion orientation of the prism. \\
                                & 4. The position and quaternion orientation of the cube. \\
                                & 5. The euclidean distance between the bottom of the prism \\ 
                                &       ~~~~     and top of the cube. \\
                                & 6. The vector between the same two points above. \\
        \hline
        \multirow{5}{*}{Selectors} & 1. The position of the pinch point between the opposed fingers. \\
                     & 2. The euclidean distance between the pinch point and the prism. \\
                     & 3. The position of the centre of mass of the prism. \\
                     & 4. The position of the centre of mass of the cube. \\
                     & 5. The euclidean distance between the bottom of the prism \\ 
                                &       ~~~~     and the top of the cube. \\

        \hline
    \end{tabular}
\end{table}

\begin{table} [htbp]
    \caption{Action vectors for each of the three concepts.}
    \label{tab:actions}
    \centering
    \begin{tabular}{l|l}
        \hline
        \textbf{Concept} & \textbf{Action} \\ 
        \hline
        \multirow{1}{*}{Orient} & Target angular velocities for the 6 joints not including the fingers. \\ & Fingers extend maximally. \\ 
        \hline
        \multirow{2}{*}{Lift} & Target angular velocities for the upper arm (1st, 2nd, and 3rd joints),  \\ & and opposed fingers (7th, and 9th joints). \\ & Remaining finger receives no command. \\
        \hline                       
        \multirow{1}{*}{Stack} & Target angular velocities for the 6 joints not including the fingers. \\ & The fingers close with moderate force. \\
        \hline
    \end{tabular}
\end{table}

\section{Conclusion}

We presented the Concept Network Reinforcement Learning (CNRL) framework, which enables true problem decomposition for reinforcement learning problems. A complex learning problem can be broken down into concepts, each concept learned independently, then reassembled into a complete solution. Decomposing problems in this way can greatly reduce the amount of training needed to achieve a useful result. 

Independent training of concepts allows each to use a focused reward function, simplifying reward design. For sub-problems where solutions already exist, such as moving a robotic arm from place to place, they can be seamlessly plugged in among learned concepts. Similarly, individual concepts can be reused as components in other tasks, or replaced with improved versions.

CNRL is suitable for industrial applications, allowing for flexible goal specifications and rapid transfer of solutions to new variants of a problem. Compared with training monolithic networks to solve complete tasks, CNRL greatly accelerates the speed with which new combinations of functionality can be trained and built upon. 

CNRL has deployment-time benefits: the training process for concepts naturally produces policies with well-defined validity regions, so they can be executed safely and reliably. It also provides improved explainability: by tracking which sub-concepts are activated when generating behavior, the system can provide context for why decisions were made.

We demonstrated CNRL on a complex robotics task requiring dexterous manipulation -- grasping a prism and precisely stacking it on a cube. We successfully solve the task, incorporating several inverse-kinematics-based classical controllers as well as a hierarchically decomposed set of learned concepts. Our approach to assembling sub-concepts into the overall solution is extremely fast, taking 45x fewer samples than a state-of-the-art approach on the same task from \cite{popov2017data}.

There are many directions for future work. We plan to tackle problems where the provided sub-concepts are not sufficient to solve the complete task, and the selector concept must synthesize additional behavior to cover the gaps. We will report on experiments applying CNRL to a wider variety of learning tasks, including tasks that require learned concepts for perception. Finally, we will apply these techniques to real-world tasks.

\bibliographystyle{model2-names}
\bibliography{refs}

\begin{thebibliography}{23}
\expandafter\ifx\csname natexlab\endcsname\relax\def\natexlab#1{#1}\fi
\providecommand{\url}[1]{\texttt{#1}}
\providecommand{\href}[2]{#2}
\providecommand{\path}[1]{#1}
\providecommand{\DOIprefix}{doi:}
\providecommand{\ArXivprefix}{arXiv:}
\providecommand{\URLprefix}{URL: }
\providecommand{\Pubmedprefix}{pmid:}
\providecommand{\doi}[1]{\href{http://dx.doi.org/#1}{\path{#1}}}
\providecommand{\Pubmed}[1]{\href{pmid:#1}{\path{#1}}}
\providecommand{\bibinfo}[2]{#2}
\ifx\xfnm\relax \def\xfnm[#1]{\unskip,\space#1}\fi
\bibitem[{Benbrahim and Franklin(1997)}]{benbrahim1997biped}
\bibinfo{author}{Benbrahim, H.}, \bibinfo{author}{Franklin, J.A.},
  \bibinfo{year}{1997}.
\newblock \bibinfo{title}{Biped dynamic walking using reinforcement learning}.
\newblock \bibinfo{journal}{Robotics and Autonomous Systems}
  \bibinfo{volume}{22}, \bibinfo{pages}{283--302}.
\bibitem[{Finn et~al.(2016)Finn, Levine and Abbeel}]{finn2016inverse}
\bibinfo{author}{Finn, C.}, \bibinfo{author}{Levine, S.},
  \bibinfo{author}{Abbeel, P.}, \bibinfo{year}{2016}.
\newblock \bibinfo{title}{Guided cost learning: Deep inverse optimal control
  via policy optimization}.
\newblock \bibinfo{journal}{arXiv:1603.00448} .
\bibitem[{Gabillon et~al.(2013)Gabillon, Ghavamzadeh and
  Scherrer}]{gabillon2013approximate}
\bibinfo{author}{Gabillon, V.}, \bibinfo{author}{Ghavamzadeh, M.},
  \bibinfo{author}{Scherrer, B.}, \bibinfo{year}{2013}.
\newblock \bibinfo{title}{Approximate dynamic programming finally performs well
  in the game of tetris}, in: \bibinfo{booktitle}{Advances in neural
  information processing systems}, pp. \bibinfo{pages}{1754--1762}.
\bibitem[{Gu et~al.(2016)Gu, Holly, Lillicrap and Levine}]{gu2016anaf}
\bibinfo{author}{Gu, S.}, \bibinfo{author}{Holly, E.},
  \bibinfo{author}{Lillicrap, T.}, \bibinfo{author}{Levine, S.},
  \bibinfo{year}{2016}.
\newblock \bibinfo{title}{Deep reinforcement learning for robotic manipulation
  with asynchronous off-policy updates}.
\newblock \bibinfo{journal}{arXiv preprint arXiv:1610.00633} .
\bibitem[{Hansen and Ostermeier(1996)}]{cma-es}
\bibinfo{author}{Hansen, N.}, \bibinfo{author}{Ostermeier, A.},
  \bibinfo{year}{1996}.
\newblock \bibinfo{title}{Adapting arbitrary normal mutation distributions in
  evolution strategies: The covariance matrix adaptation.}, in:
  \bibinfo{booktitle}{Proceedings of IEEE International Conference on}, pp.
  \bibinfo{pages}{312--317}.
\bibitem[{Kakade and Langford(2002)}]{conserve_iter}
\bibinfo{author}{Kakade, S.}, \bibinfo{author}{Langford, J.},
  \bibinfo{year}{2002}.
\newblock \bibinfo{title}{Approximately optimal approximate reinforcement
  learning}, in: \bibinfo{booktitle}{The 19st International Conference on
  Machine Learning}, pp. \bibinfo{pages}{267--274}.
\bibitem[{Kulkarni et~al.(2016)Kulkarni, Narasimhan, Saeedi and
  Tenenbaum}]{kulkarni2016hierarchical}
\bibinfo{author}{Kulkarni, T.D.}, \bibinfo{author}{Narasimhan, K.},
  \bibinfo{author}{Saeedi, A.}, \bibinfo{author}{Tenenbaum, J.},
  \bibinfo{year}{2016}.
\newblock \bibinfo{title}{Hierarchical deep reinforcement learning: Integrating
  temporal abstraction and intrinsic motivation}, in:
  \bibinfo{booktitle}{Advances in Neural Information Processing Systems}, pp.
  \bibinfo{pages}{3675--3683}.
\bibitem[{Levine et~al.(2016)Levine, Pastor, Krizhevsky and
  Quillen}]{levine2016visuo}
\bibinfo{author}{Levine, S.}, \bibinfo{author}{Pastor, P.},
  \bibinfo{author}{Krizhevsky, A.}, \bibinfo{author}{Quillen, D.},
  \bibinfo{year}{2016}.
\newblock \bibinfo{title}{Learning hand-eye coordination for robotic grasping
  with deep learning and large-scale data collection}.
\newblock \bibinfo{journal}{arXiv preprint arXiv:1603.02199} .
\bibitem[{Lillicrap et~al.(2015)Lillicrap, Hunt, Pritzel, Heess, Erez, Tassa,
  Silver and Wierstra}]{ddpg}
\bibinfo{author}{Lillicrap, T.P.}, \bibinfo{author}{Hunt, J.J.},
  \bibinfo{author}{Pritzel, A.}, \bibinfo{author}{Heess, N.},
  \bibinfo{author}{Erez, T.}, \bibinfo{author}{Tassa, Y.},
  \bibinfo{author}{Silver, D.}, \bibinfo{author}{Wierstra, D.},
  \bibinfo{year}{2015}.
\newblock \bibinfo{title}{Continuous control with deep reinforcement learning}.
\newblock \bibinfo{journal}{arXiv preprint arXiv:1509.02971} .
\bibitem[{Mannor et~al.(2003)Mannor, Rubinstein and Gat.}]{cem}
\bibinfo{author}{Mannor, S.}, \bibinfo{author}{Rubinstein, R.},
  \bibinfo{author}{Gat., Y.}, \bibinfo{year}{2003}.
\newblock \bibinfo{title}{The cross entropy method for fast policy search}, in:
  \bibinfo{booktitle}{International Conference on Machine Learning}, pp.
  \bibinfo{pages}{512--519}.
\bibitem[{Mnih et~al.(2015)Mnih, Kavukcuoglu, Silver, Rusu, Veness, Bellemare,
  Graves, Riedmiller, Fidjeland, Ostrovski et~al.}]{mnih2015human}
\bibinfo{author}{Mnih, V.}, \bibinfo{author}{Kavukcuoglu, K.},
  \bibinfo{author}{Silver, D.}, \bibinfo{author}{Rusu, A.A.},
  \bibinfo{author}{Veness, J.}, \bibinfo{author}{Bellemare, M.G.},
  \bibinfo{author}{Graves, A.}, \bibinfo{author}{Riedmiller, M.},
  \bibinfo{author}{Fidjeland, A.K.}, \bibinfo{author}{Ostrovski, G.}, et~al.,
  \bibinfo{year}{2015}.
\newblock \bibinfo{title}{Human-level control through deep reinforcement
  learning}.
\newblock \bibinfo{journal}{Nature} \bibinfo{volume}{518},
  \bibinfo{pages}{529--533}.
\bibitem[{Popov et~al.(2017)Popov, Heess, Lillicrap, Hafner, Barth-Maron,
  Vecerik, Lampe, Tassa, Erez and Riedmiller}]{popov2017data}
\bibinfo{author}{Popov, I.}, \bibinfo{author}{Heess, N.},
  \bibinfo{author}{Lillicrap, T.}, \bibinfo{author}{Hafner, R.},
  \bibinfo{author}{Barth-Maron, G.}, \bibinfo{author}{Vecerik, M.},
  \bibinfo{author}{Lampe, T.}, \bibinfo{author}{Tassa, Y.},
  \bibinfo{author}{Erez, T.}, \bibinfo{author}{Riedmiller, M.},
  \bibinfo{year}{2017}.
\newblock \bibinfo{title}{Data-efficient deep reinforcement learning for
  dexterous manipulation}.
\newblock \bibinfo{journal}{arXiv preprint arXiv:1704.03073} .
\bibitem[{Precup(2000)}]{precup2000temporal}
\bibinfo{author}{Precup, D.}, \bibinfo{year}{2000}.
\newblock \bibinfo{title}{Temporal abstraction in reinforcement learning} .
\bibitem[{Rusu et~al.(2015)Rusu, Colmenarejo, Gulcehre, Desjardins,
  Kirkpatrick, Pascanu, Mnih, Kavukcuoglu and Hadsell}]{rusu2015policy}
\bibinfo{author}{Rusu, A.A.}, \bibinfo{author}{Colmenarejo, S.G.},
  \bibinfo{author}{Gulcehre, C.}, \bibinfo{author}{Desjardins, G.},
  \bibinfo{author}{Kirkpatrick, J.}, \bibinfo{author}{Pascanu, R.},
  \bibinfo{author}{Mnih, V.}, \bibinfo{author}{Kavukcuoglu, K.},
  \bibinfo{author}{Hadsell, R.}, \bibinfo{year}{2015}.
\newblock \bibinfo{title}{Policy distillation}.
\newblock \bibinfo{journal}{arXiv preprint arXiv:1511.06295} .
\bibitem[{Schulman et~al.(2015a)Schulman, Levine, Abbeel, Jordan and
  Moritz}]{schulman2015trust}
\bibinfo{author}{Schulman, J.}, \bibinfo{author}{Levine, S.},
  \bibinfo{author}{Abbeel, P.}, \bibinfo{author}{Jordan, M.},
  \bibinfo{author}{Moritz, P.}, \bibinfo{year}{2015}a.
\newblock \bibinfo{title}{Trust region policy optimization}, in:
  \bibinfo{booktitle}{Proceedings of the 32nd International Conference on
  Machine Learning (ICML-15)}, pp. \bibinfo{pages}{1889--1897}.
\bibitem[{Schulman et~al.(2015b)Schulman, Moritz, Levine, Jordan and
  Abbeel}]{schulman2015high}
\bibinfo{author}{Schulman, J.}, \bibinfo{author}{Moritz, P.},
  \bibinfo{author}{Levine, S.}, \bibinfo{author}{Jordan, M.},
  \bibinfo{author}{Abbeel, P.}, \bibinfo{year}{2015}b.
\newblock \bibinfo{title}{High-dimensional continuous control using generalized
  advantage estimation}.
\newblock \bibinfo{journal}{arXiv preprint arXiv:1506.02438} .
\bibitem[{Silver et~al.(2016)Silver, Huang, Maddison, Guez, Sifre, Van
  Den~Driessche, Schrittwieser, Antonoglou, Panneershelvam, Lanctot
  et~al.}]{silver2016mastering}
\bibinfo{author}{Silver, D.}, \bibinfo{author}{Huang, A.},
  \bibinfo{author}{Maddison, C.J.}, \bibinfo{author}{Guez, A.},
  \bibinfo{author}{Sifre, L.}, \bibinfo{author}{Van Den~Driessche, G.},
  \bibinfo{author}{Schrittwieser, J.}, \bibinfo{author}{Antonoglou, I.},
  \bibinfo{author}{Panneershelvam, V.}, \bibinfo{author}{Lanctot, M.}, et~al.,
  \bibinfo{year}{2016}.
\newblock \bibinfo{title}{Mastering the game of go with deep neural networks
  and tree search}.
\newblock \bibinfo{journal}{Nature} \bibinfo{volume}{529},
  \bibinfo{pages}{484--489}.
\bibitem[{Silver et~al.(2014)Silver, Lever, Heess, Degris, Wierstra and
  Riedmiller}]{dpg}
\bibinfo{author}{Silver, D.}, \bibinfo{author}{Lever, G.},
  \bibinfo{author}{Heess, N.}, \bibinfo{author}{Degris, T.},
  \bibinfo{author}{Wierstra, D.}, \bibinfo{author}{Riedmiller, M.},
  \bibinfo{year}{2014}.
\newblock \bibinfo{title}{Deterministic policy gradient algorithms}, in:
  \bibinfo{booktitle}{The 31st International Conference on Machine Learning},
  pp. \bibinfo{pages}{387--395}.
\bibitem[{Spall(2003)}]{introtopg}
\bibinfo{author}{Spall, J.C.}, \bibinfo{year}{2003}.
\newblock \bibinfo{title}{Introduction to Stochastic Search and Optimization:
  Estimation, Simulation, and Control}.
\newblock \bibinfo{publisher}{Wiley-Interscience}.
\bibitem[{Sutton et~al.(1999)Sutton, Precup and Singh}]{sutton1999between}
\bibinfo{author}{Sutton, R.S.}, \bibinfo{author}{Precup, D.},
  \bibinfo{author}{Singh, S.}, \bibinfo{year}{1999}.
\newblock \bibinfo{title}{Between mdps and semi-mdps: A framework for temporal
  abstraction in reinforcement learning}.
\newblock \bibinfo{journal}{Artificial intelligence} \bibinfo{volume}{112},
  \bibinfo{pages}{181--211}.
\bibitem[{Tessler et~al.(2017)Tessler, Givony, Zahavy, Mankowitz and
  Mannor}]{tessler2017deep}
\bibinfo{author}{Tessler, C.}, \bibinfo{author}{Givony, S.},
  \bibinfo{author}{Zahavy, T.}, \bibinfo{author}{Mankowitz, D.J.},
  \bibinfo{author}{Mannor, S.}, \bibinfo{year}{2017}.
\newblock \bibinfo{title}{A deep hierarchical approach to lifelong learning in
  minecraft.}, in: \bibinfo{booktitle}{AAAI}, pp. \bibinfo{pages}{1553--1561}.
\bibitem[{Van~Hasselt et~al.(2016)Van~Hasselt, Guez and Silver}]{van2016deep}
\bibinfo{author}{Van~Hasselt, H.}, \bibinfo{author}{Guez, A.},
  \bibinfo{author}{Silver, D.}, \bibinfo{year}{2016}.
\newblock \bibinfo{title}{Deep reinforcement learning with double q-learning.},
  in: \bibinfo{booktitle}{AAAI}, pp. \bibinfo{pages}{2094--2100}.
\bibitem[{Watkins and Dayan(1992)}]{watkins1992q}
\bibinfo{author}{Watkins, C.J.}, \bibinfo{author}{Dayan, P.},
  \bibinfo{year}{1992}.
\newblock \bibinfo{title}{Q-learning}.
\newblock \bibinfo{journal}{Machine learning} \bibinfo{volume}{8},
  \bibinfo{pages}{279--292}.

\end{thebibliography}

\appendix
\section{Reward Shaping}
\label{app:shaping}
As discussed previously, the full control concept is decomposed into separate "Grasp" and "Stack" concepts (skills), while Grasp itself is decomposed into "Orient" and "Lift". In this section, we present the precise shaping reward functions used for the training of each concept.

\subsection{Orient}

\begin{equation}
    r_{\theta}=1-(0.5(\frac{min(\theta_x, \theta_y)}{45^\circ}+\frac{\theta_z}{90^\circ}))^\alpha
\end{equation}

where $r_{\theta}$ is the angular component of the shaping reward for stack and orient, $\theta_x$, $\theta_y$, and $\theta_z$ are the angle between the line passing through the two opposed fingers and the $x$, $y$, or $z$ axes in the reference frame of the target object, respectively, and $\alpha$ controls the sharpness of the shaping. Since the objects are symmetrical in $x$ and $y$, we allow any of the four orientations of the fingers that line up with the $x$ or $y$ axes by only looking at the smallest angular distance to either the $x$ or $y$ axis, yielding a distance that ranges from 0 to 45 degrees. The $z$ must uniquely line up with the object, and ranges from 0 to 90. Here, we used an $\alpha$ value of $0.4$.

\begin{equation}
    r_{d} = 1-(\frac{d}{d_{max}})^{\alpha},
\end{equation}

where $r_d$ is the shaping reward for the reaching toward the goal location, $d$ is the distance between the pinch point of the opposed fingers and the goal location, $d_{max}$ is the terminal distance for this task, and $\alpha$ controls the sharpness of the shaping.

\begin{equation}
\gamma_t = 1-\frac{t}{t_{max}}
\end{equation}

where $\gamma_t$ is a time decay factor applied to the reward to encourage fast completion, $t$ is the current time step within the episode, and $t_{max}$ is the time step limit for an episode,

\begin{equation}
R_{orient} =
\begin{cases}
\gamma_t \: b_{orient}, & \mbox{ if $d_{orient} < \epsilon_d$ \:\:and\:\: $\theta_{orient} < \epsilon_\theta$ } \\
\gamma_t \: (w_{\theta} r_{\theta} + w_{d} r_{d}), & \mbox{otherwise,} \\
\end{cases}
\end{equation}

where $R_{orient}$ is the final reward for the "Orient" concept, $b_{orient}$ is the bonus awarded on successful completion of the orient task, $d_{orient}$ is the distance between the pinch point and prism, $\theta_{orient}$ is the angle between the line connecting the opposed fingers and one of the axes of the prism, and the $\epsilon$ values for each are their tolerances. 
\subsubsection{Lift}

\begin{equation}
r_{p} = 1-(\frac{p}{p_{max}})^{\alpha}
\end{equation}

where $r_{p}$ is the pinch shaping reward component for closing the fingers, $p$ is the distance between the two opposed fingers, and $p_{max}$ is the maximum possible distance between the fingers.

\begin{equation}
r_{h} = (\frac{h}{h_{max}})^{\alpha}
\end{equation}

where $r_{h}$ is the height shaping component for lifting the prism, $h$ is the height of the prism off the ground, and $h_{max}$ is the distance at which we declare the prism lifted and terminate the episode. Here we used an $\alpha$ of $4$.

\begin{equation}
R_{\:lift} =
\begin{cases}
\gamma_t \: b_{\:lift}, & \mbox{if $h>\epsilon_h$}\\
\gamma_t \: (w_{p} +  w_{h} r_{h}), & \mbox{if $p>\epsilon_p$} \\
\gamma_t \: w_{p} r_{p}, & \mbox{otherwise,} \\
\end{cases}
\end{equation}

where $R_{\:lift}$ is the final reward for the "Lift" concept, $b_{\:lift}$ is the bonus reward assigned for successfully lifting the prism above the threshold height, h is the height of the prism, $\epsilon_h$ is the threshold height, and $\epsilon_p$ is the threshold distance between the fingers below which they are considered pinched. The bonus rewards for success are greater than the total reward that could have been accumulated had the agent remained in this highest reward state for the remaining time in the episode, to encourage fast completion of the task.

\subsection{Grasp}

Here, $R_{grasp}$ is the final shaping reward for the "Grasp" concept.

\begin{equation}
R_{grasp} =
\begin{cases}
b_{\:lift}, & \mbox{if $h>\epsilon_h$} \\
w_{\theta}, & \mbox{ if $d_{orient} < \epsilon_d$ \:\:and\:\: $\theta_{orient} < \epsilon_\theta$ } \\
0, & \mbox{otherwise,} \\
\end{cases}
\end{equation}

\subsection{Stack}

\begin{equation}
R_{stack} =
\begin{cases}
\gamma_t \: b_{stack}, & \mbox{ if $d_{stack} < \epsilon_d$ \:\:and\:\: $\theta_{stack} < \epsilon_\theta$ } \\
\gamma_t \: (w_{\theta} r_{\theta} + w_{d} r_{d}), & \mbox{otherwise,} \\
\end{cases}
\end{equation}

where $d_{stack}$ is the distance between the pinch point and the goal, $\epsilon_d$ and $\epsilon_\theta$ are the thresholds for success in angular and euclidean distance, $w_{\theta}$ and $w_{d}$ are the weights assigned to those reward components, and $b_{stack}$ is the bonus reward assigned for successful completion of the stack task.

\subsection{Full Task}

\begin{equation}
R_{full} = 
\begin{cases}
w_{stack}, & \text{if } d_{stack} < \epsilon_d \land \theta_{stack} < \epsilon_\theta \\
w_{stage_2}, & \text{if } d_{stack} < \epsilon_d \land \theta_{stack} < \epsilon_\theta \\
w_{grasp}, & \text{if } h > \epsilon_h \\
w_{stage_1}, & \text{if } d_{stage_1} < \epsilon_d \\
0, & \text{otherwise} \\
\end{cases}
\end{equation}

\section{Terminal Conditions}
\label{app:terminal}

\paragraph{Orient} For the orient concept, an episode would end early if the hand moved too far from the prism, if the prism tipped more than 15 degrees, or the goal was achieved by aligning the opposed fingers with the prism while the pinch point was 1.5cm above the prism.

\paragraph{Lift} For the lift concept, an episode would end early if the prism was moved outside a virtual cylinder centered on the starting position of the prism, if the hand moved more than a certain distance from the prism, or if the goal was achieved by lifting the prism above a target height.

\paragraph{Stack} For the stack concept, an episode would end early if the prism moved too far from the cube, if the prism touched the ground, or if the goal was achieved by lining the prism up with the cube and bringing them into contact.

\end{document}